# Does Attention Mechanism Possess the Feature of Human Reading? A Perspective of Sentiment Classification Task

Lei Zhao, Yingyi Zhang, Chengzhi Zhang[*]

Department of Information Management, Nanjing University of Science and Technology, Nanjing, China

**Abstract**

**Purpose –** To understand the meaning of a sentence, humans can focus on important words in the sentence, which reflects our eyes staying on each word in different gaze time or times. Thus, some studies utilize eye-tracking values to optimize the attention mechanism in deep learning models. But these studies lack to explain the rationality of this approach. Whether the attention mechanism possesses this feature of human reading needs to be explored.

**Design/methodology/approach –** We conducted experiments on a sentiment classification task. Firstly, we obtained eye-tracking values from two open-source eye-tracking corpora to describe the feature of human reading. Then, the machine attention values of each sentence were learned from a sentiment classification model. Finally, a comparison was conducted to analyze machine attention values and eye-tracking values.

**Findings –** Through experiments, we found the attention mechanism can focus on important words, such as adjectives, adverbs, and sentiment words, which are valuable for judging the sentiment of sentences on the sentiment classification task. It possesses the feature of human reading, focusing on important words in sentences when reading. Due to the insufficient learning of the attention mechanism, some words are wrongly focused. The eye-tracking values can help the attention mechanism correct this error and improve the model performance.

**Originality/value –** Our research not only provides a reasonable explanation for the study of using eye-tracking values to optimize the attention mechanism, but also provides new inspiration for the interpretability of attention mechanism.

**Paper type** Research paper

## 1. Introduction

The rise of artificial intelligence (AI) has profoundly changed the way humans understand the world. Machines can make a series of responses similar to humans, becoming more intelligent and even far superior to humans. The AlphaGo defeating the world champion of Go is the best example. However, in the field of natural language processing (NLP), due to the complexity and diversity of human languages, machines cannot fully understand human expressions in some tasks, e.g., machine translation (Läubli *et al.*, 2020), summary generation (Sheela and Janet, 2021), keyphrase extraction (Zhang, Zhao, *et al.*, 2022) and so on. Whether machines really possess a human-like way of thinking is worthy of in-depth exploration in this background.

When we are reading a sentence, we can focus on some words that are useful for understanding the sentence. In other words, we do not pay the same attention to all words, which reflects our eyes staying on each word in different gaze time or times. The Eye-tracking corpus (ETC) is a record of this feature of human reading. It uses eye-trackers to capture the time or fixation times of eye gaze on each word (Rayner *et al.*, 2012). ETCs have been applied to various NLP tasks, e.g., part-of-speech tagging (Barrett *et al.*, 2016), sentiment analysis (Mishra, Kanojia, Nagar, *et al.*, 2016), multiword expression (Rohanian *et al.*, 2017), keyphrase extraction (Zhang and Zhang, 2021), etc. These studies used eye-tracking values to optimize the attention mechanism, but they lack to explain the rationality of this approach. Therefore, it is essential to compare the machine attention values obtained by the model based on attention mechanism and the eye-tracking values, providing theoretical support for these studies.

The study is in the line of interpretability research of the attention mechanism. Previous research had explored the relationship between machine attention and human attention, in which human attention is represented by the important degree of words manually annotated. Generally, the important degree is in a binary distribution, represented by 0-1 (Sen *et al.*, 2020; Vashishth *et al.*, 2019), different from the continuous distribution of machine attention values. Since the binary distribution can only indicate which words are important in a sentence, but cannot compare the important degree between words, the analysis based on this cannot fully describe the relationship between machine attention and human attention. In contrast, eye-tracking values are a continuous distribution, and a larger eye-tracking value indicates that the word is more important in a sentence. Fig. 1 shows the attention distribution of machine attention values and eye-tracking values on a movie review randomly selected from ZuCo. In this example, both attention mechanism and human focus on the word "mesmerizing" in the sentence, which greatly affects judging the sentiment of this review as positive. This example shows a certain relationship between the machine attention values and the eye-tracking values.

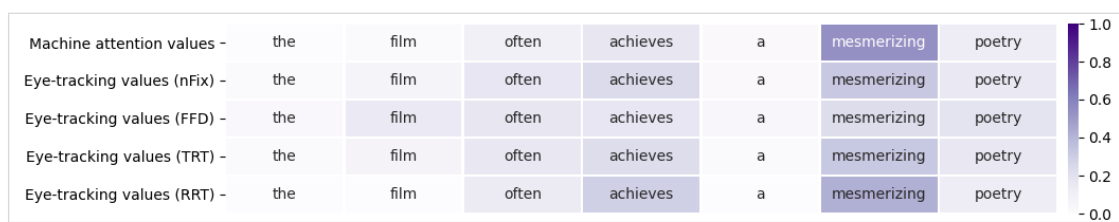

(**Note:** There are four types of eye-tracking values, i.e., nFix, FFD, TRT, and RRT, introduced in Section 3.1. The darker the color, the higher attention to the word.)

Figure 1. An example of the attention distribution of machine attention values and eye-tracking values on a movie review.

The purpose of our study is to analyze whether the attention mechanism possesses the feature of human reading, focusing on the important words in sentences when reading. We conducted experiments on a sentiment classification task. The eye-tracking values were used to describe the feature of human reading, which are collected from two open-source ETCs, i.e., ZuCo (Hollenstein *et al.*, 2018) and Eye-tracking and Sentiment Analysis-II (Mishra, Kanojia and Bhattacharyya, 2016). They are sentiment classification task-driven corpora. The machine attention values for the two ETCs were learned from a sentiment classification model based on the attention mechanism. To compare the machine attention values and the eye-tracking values, we used Spearman correlation coefficient (Maritz, 1995), Jensen Shannon divergence (Lin, 1991) to analyze the correlation between them. Combining the characteristics of the sentiment classification task, we explored whether the attention mechanism and humans focus on the same words, and calculated the attention rate of specific part-of-speech words and sentiment words. Through experiments, we found the attention mechanism can focus on important words, such as adjectives, adverbs, and sentiment words, which are valuable for judging the sentiment of sentences on the sentiment classification task. It possesses the feature of human reading, focusing on important words in a sentence when reading. Due to the insufficient learning of the attention mechanism, some words are wrongly focused. The eye-tracking values can help the attention mechanism correct this error and improve the performance of the model.

The contribution of the paper is three-fold.
1) We found that on the task of sentiment classification, the attention mechanism can focus on the important words in sentences through the learning of samples, and give greater weights. It possesses the feature of human reading, focusing on the important words in sentences. This provides new inspiration for the interpretability of attention mechanism.
2) The attention mechanism can make mistakes when assigning weights to words in a sentence. The eye-tracking values can help the attention mechanism correct this error and improve the model performance. This provides a reasonable explanation for the study of using eye-tracking values to optimize the attention mechanism.
3) We proposed to use eye-tracking values to explore the relationship between machine attention and human attention. This is different from previous work that required human attention to be collected for this study. The eye-tracking values can be a by-product of the sentiment annotation task. When annotating a sentence, the eye-tracking values can be collected at the same time.

In the rest of this paper, Section 2 summarizes the application of eye-tracking values in NLP tasks and the interpretability research of the attention mechanism. In Section 3, we describe the datasets, model structure and metrics used in the experiment. Section 4 analyzes the influencing factors of machine attention values and the comparison of them with the eye-tracking values. Sections 5 and 6 are discussion and summary.

## 2. Literature Review

Eye-tracking values reflect human attention and have been widely used in various deep learning models. However, few studies use eye-tracking values to explain the role of the attention mechanism. This section mainly summarizes the application of eye-tracking values in NLP tasks and the interpretability research of the attention mechanism.

### 2.1 Application of Eye-tracking Values in NLP Tasks

In recent years, eye-tracking values, including total fixation time, fixation times, and first fixation time, have been used for part-of-speech tagging (Barrett *et al.*, 2016), syntactic analysis (Agrawal and Rosa, 2020), sentence classification (Barrett *et al.*, 2018; Long *et al.*, 2017), information extraction (Barrett and Hollenstein, 2020; Hollenstein and Zhang, 2019; Zhang and Zhang, 2021), and text understanding (Malmaud *et al.*, 2020; Zheng *et al.*, 2019), which has been proved to improve the performance of deep learning models. Besides, eye-tracking values are also widely used in behavioral sciences to analyze users' behaviors, such as searching and querying information (Gwizdka *et al.*, 2019; Sachse, 2019; Wu and Huang, 2018).

Eye-tracking values are generally obtained from open-source ETCs, including ZuCo, DUNDEE, and GECO. At present, there are four methods to introduce eye-tracking values into deep learning models. The first reduces the difference between machine attention values and eye-tracking values through training. For example, Zhang and Zhang (2021) proposed a keyphrase extraction model and used eye-tracking values as ground truth to optimize machine attention values. The second uses eye-tracking values as machine attention values, or uses multiplication or addition to combine eye-tracking values with machine attention values. For example, Long *et al.* (2017) proposed a cognitive-based attention model. In this model, they first used eye-tracking values and local text features to establish a simple and effective regression model to predict reading time, and then used the predicted reading time to construct the attention layer in the neuro-sentiment analysis model. The third is to convert eye-tracking values into feature vectors and input them into models. For example, Hollenstein and Zhang (2019) embedded eye-tracking values into the vectorized representation of words for named entity recognition (NER) tasks. They found that the enhanced model was better than the baseline model. The last is to add an eye-tracking prediction module to deep learning models. For example, Sood *et al.* (2020) proposed a hybrid text saliency model (TSM), which can combine a cognitive model of reading with explicit human gaze supervision in a single machine learning framework. Besides, they further proposed a joint modeling approach to integrate TSM predictions into the attention layer of a network designed for a specific upstream NLP task without the need for any task-specific human gaze data.

For the four methods mentioned above, the first method and the second method aim to use eye-tracking values to optimize machine attention values. However, before using these methods, it is necessary to compare machine attention values and eye-tracking values to analyze whether machine attention is consistent with human attention, but previous studies lacked such analysis.

**2.2 Interpretability Research of Attention Mechanism**

Attention mechanism was first used in machine translation tasks (Bahdanau *et al.*, 2015). Subsequently, since it helps to improve the performance of the model, it is widely used in various NLP tasks, such as text classification (Li *et al.*, 2019; Liu and Guo, 2019), machine translation (Britz *et al.*, 2017; Vaswani *et al.*, 2017), recommendation (Wang *et al.*, 2018) and so on.

In recent years, the interpretability research of attention mechanism has attracted the attention of scholars (Jain and Wallace, 2019; Lei, 2017; Mullenbach *et al.*, 2018; Sen *et al.*, 2020; Wiegreffe and Pinter, 2020). For example, some studies focus on analyzing what self-attention and multi-head self-attention learn in pre-trained language models (Clark *et al.*, 2019; Goldberg, 2019; Hoover *et al.*, 2020; Htut *et al.*, 2019). Other studies focus on analyzing what traditional attention mechanisms, such as additive attention learns (Jain and Wallace, 2019; Sen *et al.*, 2020; Wiegreffe and Pinter, 2020). By analyzing the methods they use, they can be roughly divided into two ways. The first is to change the attention value itself, such as replacing the highest value or randomly generating the attention value. The purpose is to analyze whether these changes affect the performance of the model, so as to analyze the usefulness of attention mechanism. For example, Jain and Wallace(2019) found that the learned attention values are usually not associated with the gradient-based feature importance measurement. They also found that very different attention values can bring the same prediction. Therefore, they believe that attention values basically cannot improve the interpretability of the model. Based on this research, Wiegreffe and Pinter (2019) refuted that the effect of fixed attention values are as good as the learned attention values on some datasets. They concluded that random values or anti-disturbance values cannot prove that attention mechanism cannot be used as an explanation for the prediction results. The second is to analyze whether machine attention pays attention to important parts by comparing human attention. In this way, the importance of words is generally annotated manually to quantify human attention. Then, by analyzing the correlation between machine attention and human attention, explore whether attention mechanism has the ability to simulate humans (Sen *et al.*, 2020; Zhang, Sen, *et al.*, 2022).

Based on the previous studies, we proposed to use eye-tracking values to explore the relationship between machine attention and human attention. The eye-tracking values are a record of human gaze times and reading time on each word, which can reflect the feature of human reading, focusing on the important words in sentences. The more gaze times or reading time, the higher human attention on words. By comparing the machine attention values obtained by the model based on attention mechanism and the eye-tracking values, we explore whether the attention mechanism possessses this feature of human reading.

## 3. Methodology

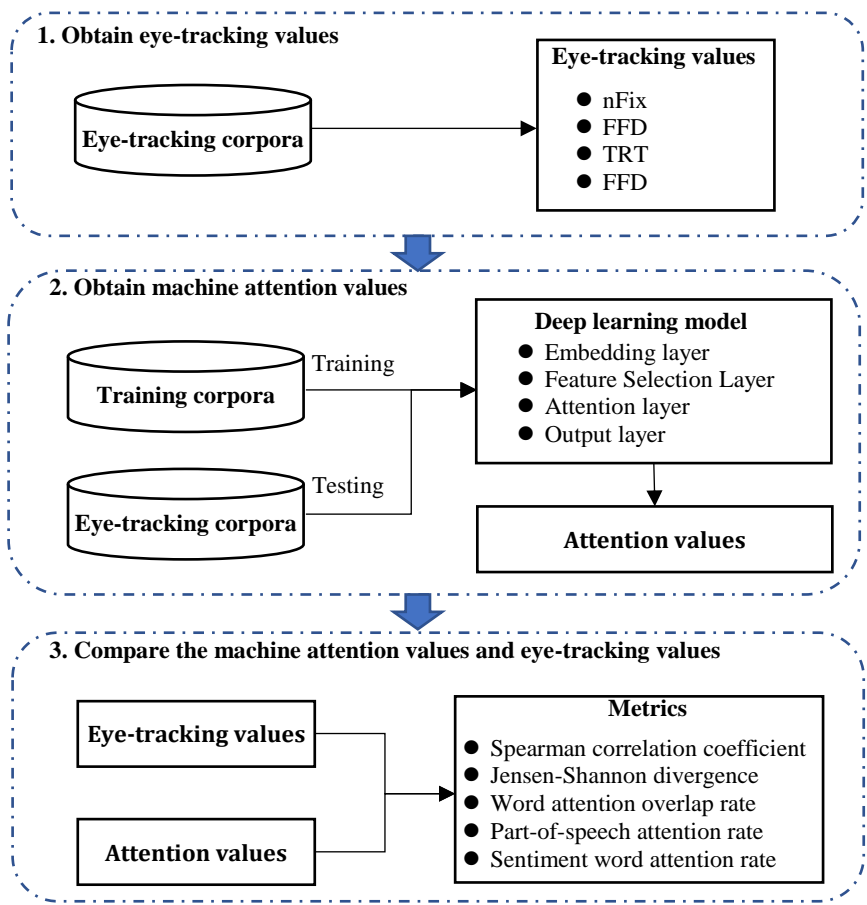

Figure 2. Research Framework

To answer whether the attention mechanism possesses the feature of human reading, focusing on the important words in sentences when reading, we need to compare the machine attention values obtained by the model based on attention mechanism and the eye-tracking values. Specifically, we first obtain eye-tracking values from two open-source ETCs, which are a record of this feature of human reading. Then, the machine attention values for the two ETCs are learned from a sentiment classification model based on attention mechanism. Finally, five metrics are used to compare the machine attention values and the eye-tracking values. The framework of this study is shown in Fig. 2. In this section, we mainly introduce the two open-source ETCs, model structure, and five metrics used in the experiment.

**3.1 Eye-tracking Corpora**

Two open-source ETCs are used for obtaining eye-tracking values, i.e., task 1 subset from ZuCo (Hollenstein *et al.*, 2018) and Eye-tracking and Sentiment Analysis-II (Mishra, Kanojia and Bhattacharyya, 2016). They are sentiment classification task-driven corpora, containing four eye-tracking values. Table 1 describes the specific meanings of four types of eye-tracking values. The following is a brief introduction to the two ETCs.

Table 1. Description of four types of eye-tracking values

| Name | Unit of measure | Description |
| --- | --- | --- |
| nFix | times | The total number of fixation times on the current word |
| FFD | ms | The duration of the first fixation on the current word |
| TRT | ms | The sum of all fixation durations on the current word |
| RRT | ms | The lookback duration for the current word, equal to TRT – FFD |

**Task 1 subdataset from ZuCo (ZuCo):** The corpus consists of 400 sentences randomly selected from the movie reviews of Stanford sentiment treebank corpus (Socher *et al.*, 2013), with three manually labeled sentiment tags, i.e., positive, neutral, and negative. Among them, 123 sentences are neutral, 137 sentences are negative, and 140 sentences are positive. In the experiment, 12 subjects were presented with positive, negative or neutral sentences for normal reading. As the subjects read the sentences, the subjects' eye-tracking values were recorded. To control the quality of their reading behavior, the subjects had to rate the quality of the described movies in 47 of the 400 sentences.

**Eye-tracking and Sentiment Analysis-II (ETSA-II):** There are 392 sentences in the corpus, of which 247 sentences are labeled as a negative tag, and 145 sentences are labeled as a positive tag. In the experiment, 7 subjects were required to read one sentence at a time and annotate it with binary labels (i.e., positive and negative). To control the reading quality, the subjects must carefully read the sentences and provide correct annotations as much as possible. Meanwhile, the eye-tracking values of subjects were recorded. Different from the ZuCo corpus, the sentences in this corpus are selected from two satirical citation websites, i.e., the Amazon movie corpus and Twitter posts, but the process of recorded eye-tracking values is similar. Both include four types of eye-tracking values, i.e., nfix, FFD, TRT, and RRT, and the measurement units are the same (see Table 1).

In the two ETCs, the same sentence has eye-tracking records of multiple subjects. Therefore, we performed summation and normalization operations to facilitate subsequent experiments. Table 2 shows an example of four eye-tracking values after preprocessing from ZuCo.

Table 2. An example of four eye-tracking values after preprocessing from ZuCo

| Word Type | The | film | Often | achieves | a | mesmerizing | poetry. |
|---|---|---|---|---|---|---|---|
| nFix | 0.02 | 0.06 | 0.17 | 0.24 | 0.03 | 0.32 | 0.15 |
| FFD | 0.04 | 0.15 | 0.18 | 0.17 | 0.05 | 0.23 | 0.19 |
| TRT | 0.02 | 0.07 | 0.16 | 0.23 | 0.02 | 0.33 | 0.16 |
| RRT | 0.00 | 0.00 | 0.15 | 0.29 | 0.00 | 0.43 | 0.13 |

**3.2 Sentiment Classification Model and Training Dataset**

To learn the machine attention values for ZuCo and ETSA-II, we constructed a sentiment classification model based on attention mechanism. This section introduces the construction of the sentiment classification model based on attention mechanism and the dataset used for model training.

**3.2.1 Structure of Sentiment Classification Model Based on Attention Mechanism**

The attention mechanism produces a non-negative distribution with a sum of 1 based on the input representation (such as word embedding vector). Then, the distribution weights the original input to obtain contextual representations as input to downstream modules (Bahdanau *et al.*, 2015). In NLP tasks, commonly used attention mechanisms are additive attention (Bahdanau *et al.*, 2015) and self attention (Vaswani *et al.*, 2017). The output of self attention is a word-to-word attention matrix, which is different from the one-dimensional vector representation of eye-tracking values. Therefore, we used the additive attention to construct a sentiment classification model, which calculated the attention value of each word in a sentence. The calculation formula of attention value is as follows:

$$\alpha_i = \frac{\exp(s(x_i))}{\Sigma_{j=i}^{N}\exp(s(x_i))} \quad (1)$$

Where $\alpha_i$ is the attention value that appears at the i-th position in a sentence, $x_i$ is the vector representation that appears at the i-th position in a sentence. $s(x_i)$ is the attention scoring function, and *N* is the number of words in a sentence. In order to reduce the learning parameters, we simplified the scoring function calculation formula, as follows:

$$s(x_i) = V^T * Tanh(x_i) \quad (2)$$

Where vector *V* is a learnable model parameter.

Fig. 3 is the framework of the sentiment classification model based on attention mechanism. We use three word vectors, i.e., Embedding, GloVe (Pennington *et al.*, 2014), and fastText (Joulin *et al.*, 2017), to train the model. The Embedding word vector is randomly initialized, which needs the model to adjust its value after learning. The GloVe[①] and fastText[②] pre-training word vectors are derived from the officially trained word vectors. For some out-of-vocabulary words in GloVe and fastText, we fill them with zero vectors, and optimize their vectorial representation through model learning. Besides, based on the best word vector, LSTM (long short term memory) (Hochreiter and Schmidhuber, 1997) and BiLSTM (bi-directional long short term memory) (Graves and Schmidhuber, 2005) are added to the model as coding layers. We also consider obtaining attention values in a clean environment without using any coding layer (None). Above the coding layer is the attention layer. We use formulas (1) and (2) to calculate the attention value of each word $\{\alpha_1, \alpha_2, ..., \alpha_{n-1}, \alpha_n\}$ in a sentence $\{w_1, w_2, ..., w_{n-1}, w_n\}$, so as to be used for subsequent analysis. The last layer is the dense layer, which outputs the sentiment tag of the input text. It's worth noting that we do not employ pre-trained language model such as BERT (Devlin *et al.*, 2019) as codding

---

[①] https://nlp.stanford.edu/data/glove.42B.300d.zip
[②] https://dl.fbaipublicfiles.com/fasttext/vectors-english/crawl-300d-2M.vec.zip

layer because a word will be split into multiple tokens, which makes it difficult to calculate the attention value of each word.

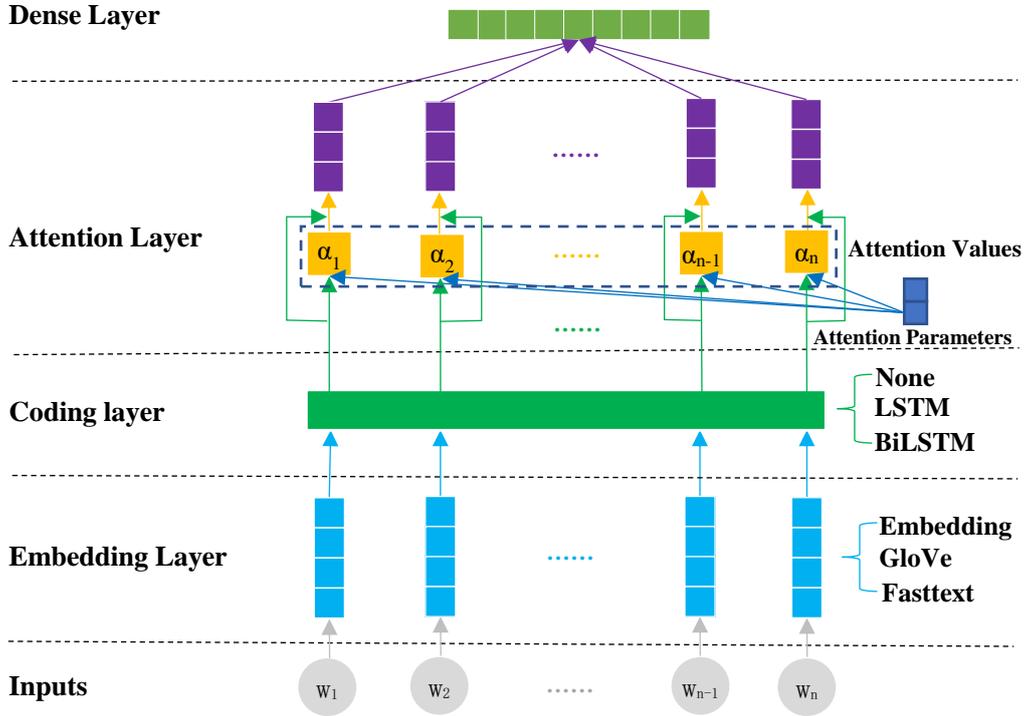

**Note:** $w_i$ represents the word that appears at the i-th position in a sentence. $\alpha_i$ represents the attention value at the i-th position in a sentence.

Figure 3. The framework of sentiment classification model based on additive attention

### 3.2.2 Construction of Model Training Dataset

We use the Stanford sentiment treebank corpus (Socher *et al.*, 2013) contains movie reviews as the training dataset for the sentiment classification model. The reasons for choosing this dataset are that ZuCo is a subset of it, and ETSA-II also consists of movie reviews (see Section 3.1). This corpus contains 11,855 sentences, and each sentence has a sentiment score. The value range is 0-1. If the value approaches 1, it's biased towards positive sentiment, and vice versa.

Since ZuCo and ETSA-II have three and two sentiment categories, respectively, we construct two training datasets with different sentiment tags based on sentiment score. Table 3 shows the number of samples in the training and test datasets corresponding to ZuCo and ETSA-II and the sentiment score interval of each category. It is worth noting that the test datasets are the two eye-tracking corpora themselves. In addition, in the ZuCo training dataset, we removed the same samples as the ZuCo test dataset.

Table 3. Number of samples used for model training and testing

| Dataset<br>Sentiment | ZuCo | | ETSA-II | |
|---|---|---|---|---|
| | **Train** | **Test** | **Train** | **Test** |
| Negative / [0.0,0.4] | 4446 | 137 | 4570 | 247 |
| Neutral / (0.4,0.6] | 1990 | 123 | - | - |
| Positive / (0.6,1.0] | 4858 | 140 | 5001 | 145 |
| **Total** | **11294** | **400** | **9571** | **392** |

## 3.3 Measurement between Machine Attention Values and Eye-tracking Values

Machine attention values and eye-tracking values are continuous distribution and don't follow any specific probabilistic distribution. According to this characteristic, we use five metrics to measure machine attention values and eye-tracking values, i.e., Spearman correlation coefficient (Maritz, 1995), Jensen Shannon divergence (Lin, 1991), word attention overlap rate, part-of-speech attention rate, and sentiment word attention rate. The first two are used to investigate correlations. The last three are self-designed metrics for this study and are described in this section.

### 3.3.1 Spearman Correlation Coefficient (SCC)

Spearman correlation coefficient is derived from the concept of product-moment correlation by Spearman (Maritz, 1995). It is mainly used to solve the problems related to name data and order data. The calculation formula is shown in formula (3).

$$SCC = 1 - \frac{6\sum(d_i^2)}{n(n^2-1)} \quad (3)$$

The calculation process is as follows. First, sort the data of two variables $(x, y)$. Then, write down the position $(x', y')$ after sorting, and the value of $(x', y')$ is called rank. The difference of rank is $d_i$ in formula (1), and n is the number of components in the variable. Finally, bring it into the formula to get the result.

### 3.3.2 Jensen Shannon Divergence (JSD)

Kullback Leibler divergence (KL divergence) was proposed by kullback and Leibler in 1951 (Kullback and Leibler, 1951). It is mainly used to measure the distance between two variables from the perspective of a probability distribution. The calculation formula is shown in formula (4). Because the KL divergence is asymmetric, Lin (1991) proposed a variant - Jensen Shannon divergence. The calculation formula is shown in formula (5).

$$KL(P||Q) = \sum_{i=1}^{n} P_i \log(\frac{P_i}{Q_i}) \quad (4)$$

$$JSD(P||Q) = \frac{1}{2}KL(P||\frac{P+Q}{2}) + \frac{1}{2}KL(Q||\frac{P+Q}{2}) \quad (5)$$

Where $P$ and $Q$ are the distributions of two variables, and n is the number of components in $P$ and $Q$.

### 3.3.3 Word Attention Overlap Rate (WAOR)

To measure whether the attention mechanism and humans focus on the same words in the sentence, we designed a new metric - word attention overlap rate. Given two attention distributions $P$ and $Q$, and a word window of $N$, the formula is:

$$WAOR = \frac{set(sort(P,\text{DESC}),N) \cap set(sort(Q,\text{DESC}),N)}{N} \quad (6)$$

Where $sort(*, DESC)$ is a sorting function, which is used to arrange words in descending order according to values, and $set(*, N)$ is an interception function, which is used to intercept the first $N$ words from the word list. The value range of WAOR is 0-1. When the value approaches 0, $P$ and $Q$ are not similar, and vice versa, and vice versa.

### 3.3.4 Part-of-speech Attention Rate (POSAR)

The part-of-speech attention rate calculates the attention degree to a part-of-speech in a word window. Given a sentence $S$ and the corresponding attention distribution $D$, on the premise that the word window is n, the formula of attention rate to part-of-speech p is:

$$POSAR_p = \frac{rate(set(sort(D,DESC),N),\ p)}{rate(S,p)} \tag{7}$$

Where $sort(*,DESC)$ is a descending sorting function, and $set(*,N)$ is an interception function, which is the same as the function in formula (6). $rate(*,p)$ is a proportional function to calculate the proportion of part-of-speech p in a word list or sentence. Its value range is greater than 0. When the value is between (0, 1), the POSAR of p is low. When the value is greater than 1, the POSAR of p is high, and the larger the value, the higher the attention.

### 3.3.5 Sentiment Word Attention Rate (SWAR)

When we judge the sentiment of a sentence, we usually focus on the sentiment words. To quantify it, we proposed the sentiment word attention rate. Given a sentence $S$ and the corresponding attention distribution $D$, we first need to label the sentiment words according to a sentiment dictionary. Here we use HowNet sentiment dictionary[③]. Assuming that there are m sentiment words in sentence $S$, the formula is:

$$SWAR = \frac{sw(set(sort(D,DESC),m))}{m} \tag{8}$$

Where $sort(*,DESC)$ is a descending sorting function, and $set(*,N)$ is an interception function, which is the same as the function in formula (6). $sw(*)$ counts the number of sentiment words in the word list. Its value range is 0-1. When the value approaches 1, the higher attention to sentiment words, and vice versa.

The above metrics are calculated for a single sentence, but different corpora contain multiple sentences. For comparison, we first calculate the metric value of each sentence, and then calculate the average value. For Spearman correlation coefficient, we only pay attention to correlation value. Therefore, when calculating this metric, we calculate the absolute value.

## 4. Experiment and Result Analysis

We conducted experiments on ZuCo and ETSA-II corpora. First, we analyzed the performance of the sentiment classification model based on additive attention. Then, we analyzed the effect of adding LSTM and BiLSTM to the model on machine attention values. Besides, the model also made mistakes in judging the sentiment of sentences. In order to avoid the influence of wrong prediction on subsequent experiments, the difference of machine attention values under correct and wrong prediction is analyzed. Next, a comparison was conducted to analyze machine attention values and eye-tracking values. Finally, we used the eye-tracking values to fine-tune the machine attention values in the model, and observe the effect of the eye-tracking values on the model performance and the machine attention values.

### 4.1 Performance of Sentiment Classification Model

The primary concern is whether our sentiment classification model has good learning ability and whether the learned machine attention values can be used for subsequent analysis. Therefore, on the

---
[③] https://github.com/thunlp/OpenHowNet/blob/master/HowNet_Dict.zip

two test datasets of ZuCo and ETSA-II, we used the macro average of accuracy, recall, and $F_1$ value to evaluate the performance of the sentiment classification model without or with the attention mechanism. To compare the performance between models, we used Support Vector Machine (SVM) (Vapnik and Chervonenkis, 1964) and BERT (Devlin *et al.*, 2019) as baseline models. The results are shown in Table 4.

It can be seen from Table 4 that the performance of the models on ZuCo test dataset is generally low, and the $F_1$ value of BERT model can only reach 72.54%, while the performance of the models on ETSA-II test dataset is better. This may be caused by the different difficulties of the two tasks. Compared with randomly initialized word vector Embedding, the use of pre-trained word vectors GloVe and fastText can improve the performance of the models. On the ZuCo and ETSA-II test datasets, the models using fastText and GloVe have the best performance, respectively. Adding LSTM and BiLSTM to the models can further improve the performance, especially on the ETSA-II test dataset. Introducing attention mechanism can improve the performance of the models. Specifically, the $F_1$ value of the models has increased, with a minimum increase of 0.1% and a maximum increase of 3.59%.

Through analysis, we found that our sentiment classification model has good learning ability. This further shows that our model can learn the relative importance of words in a sentence, and the machine attention values obtained on this model are valuable for subsequent analysis.

Table 4. The performance of sentiment classification models (%)

| Baseline Model | ZuCo | | | ETSA-II | | |
|---|---|---|---|---|---|---|
| | P | R | $F_1$ | P | R | $F_1$ |
| SVM | 55.33 | 54.56 | 54.59 | 68.45 | 68.18 | 68.30 |
| BERT | 73.46 | 74.10 | 72.54 | 84.56 | 86.75 | 84.99 |

| Experimental model | W/O | | | W | | | W/O | | | W | | |
|---|---|---|---|---|---|---|---|---|---|---|---|---|
| | P | R | $F_1$ | P | R | $F_1$ | P | R | $F_1$ | P | R | $F_1$ |
| Embedding_None | 66.27 | 66.04 | 63.13 | 63.18 | 63.75 | 61.31 | 71.96 | 72.97 | 72.26 | 75.51 | 76.88 | 75.85 |
| GloVe_None | 66.65 | 66.83 | 64.04 | 69.33 | 67.33 | 64.30 | 77.05 | 77.87 | 77.38 | **78.68** | **80.31** | **79.08** |
| FastText_None | 67.60 | 67.03 | 64.63 | **68.18** | **67.89** | **65.92** | 79.16 | 79.77 | 79.43 | 77.65 | 79.44 | 77.91 |
| BW_LSTM | 67.08 | 67.72 | 66.91 | 67.13 | 66.98 | 67.01 | 79.72 | 81.87 | 79.47 | 80.86 | 82.27 | 81.33 |
| BW_BiLSTM | 67.37 | 67.61 | 67.27 | 67.49 | 68.30 | 67.55 | 80.30 | 82.09 | 80.71 | 81.79 | 83.59 | 82.26 |

**Note**: "W/O" means that the model doesn't add attention mechanism, "W" means that the model adds attention mechanism, and "BW" means that a word vector that makes the model perform best. Specifically, on the ZuCo and ETSA-II test datasets, the models using fastText and GloVe have the best performance, respectively.

**4.2 Factors Affecting Machine Attention Values**

In section 4.1, we found that adding LSTM and BiLSTM can improve the performance of the models. Does adding LSTM and BiLSTM to the models affect machine attention? After adding LSTM and BiLSTM, can the machine attention values obtained from the models be used for subsequent analysis? The models may make mistakes in judging the sentiment of sentences, so is there a difference of machine attention values under correct and wrong prediction? This section mainly answers these questions.

**4.2.1 Influence of LSTM and BiLSTM on Machine Attention Values**

For the same sentence, different models can generate different machine attention values. To study whether adding LSTM and BiLSTM to the models can affect the machine attention values, we calculated six statistics describing the distribution characteristics, i.e., range (R), interquartile distance (IQR), standard deviation (STD), coefficient of variation (CV), skew and kurt. One-way ANOVA is used to test whether there is a significant difference in the machine attention values obtained with and without LSTM, BiLSTM. The results are shown in Fig. 4.

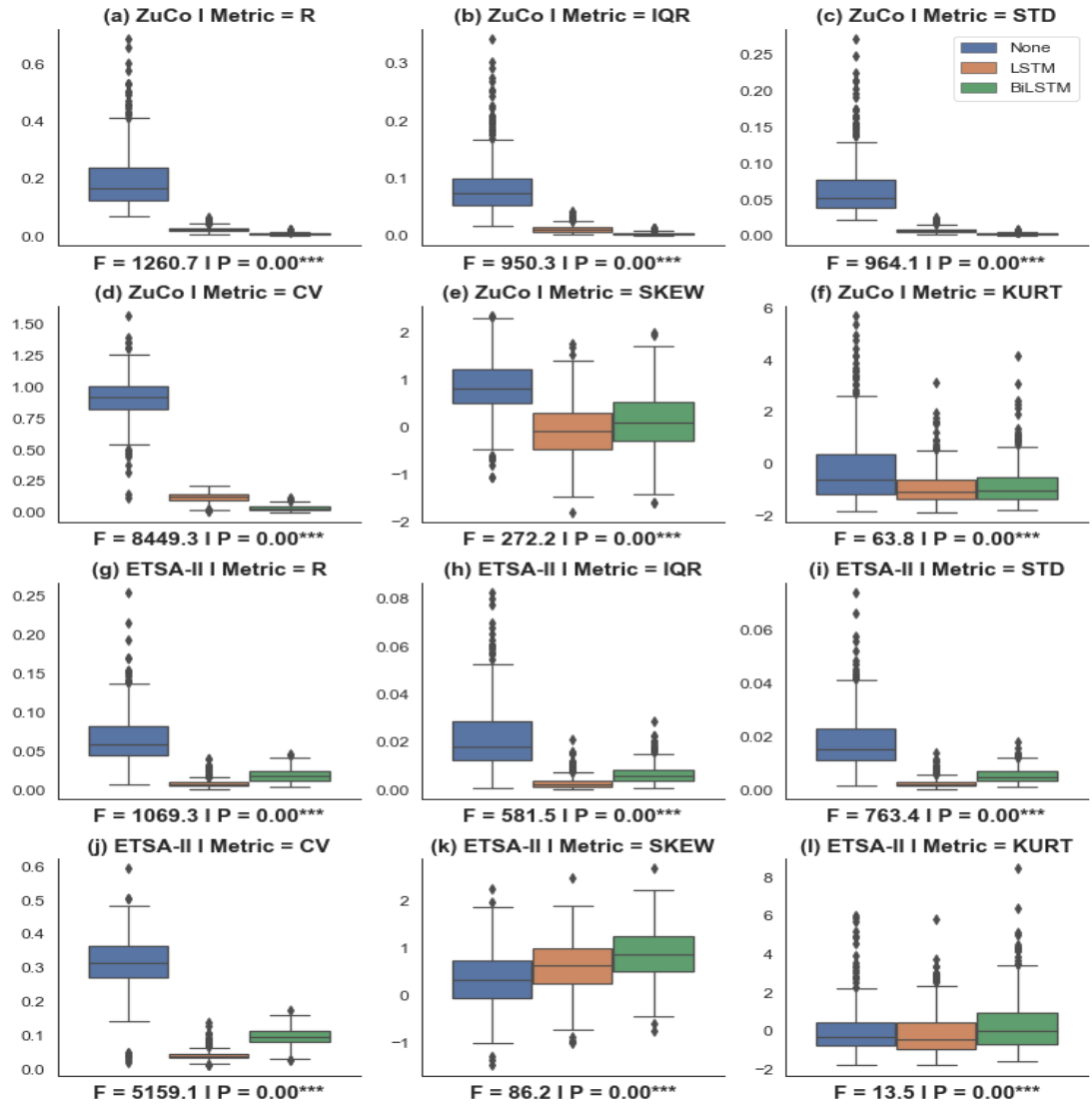

Note: ***. Significant at $\alpha$=0.001.

Figure 4. The one-way ANOVA test results of machine attention values obtained with and without LSTM and BiLSTM

***There is a significant differences in the machine attention values obtained with and without LSTM, BiLSTM.*** As shown from Fig. 4 (a) ~ (l), when the significance level α is 0.001, the p-values of all statistics are less than 0.001. The results show significant differences in the machine attention values obtained with and without LSTM, BiLSTM in the models. In particular, the machine attention values obtained without adding LSTM and BiLSTM is significantly different from the other two cases in six statistics. It further shows that adding LSTM and BiLSTM impacts the machine attention values.

Sen *et al.* (2020) pointed out that bi-Directional RNNs with additive attention demonstrate strong similarities to human attention, while uni-directional RNNs with attention differ from human attention significantly. This indicates that the RNNs can affect the machine attention. LSTMs is a variant of RNNs. Therefore, we used the attention values obtained from the model without any coding layer for the analysis between the attention values and eye-tracking values. The aim is to analyze how the attention mechanism performs in a clean environment. Specifically, the machine attention values obtained by the fastText_None model and the GloVe_None model are used on ZuCo and ETSA-II, respectively.

### 4.2.2 Influence of Prediction Results on Machine Attention Values

Since the performance of the sentiment classification model is limited, the sentiment tag of each review cannot be predicted correctly. It is necessary to consider whether there is a difference between the machine attention values under the correct and wrong prediction results, which may affect the subsequent experimental analysis.

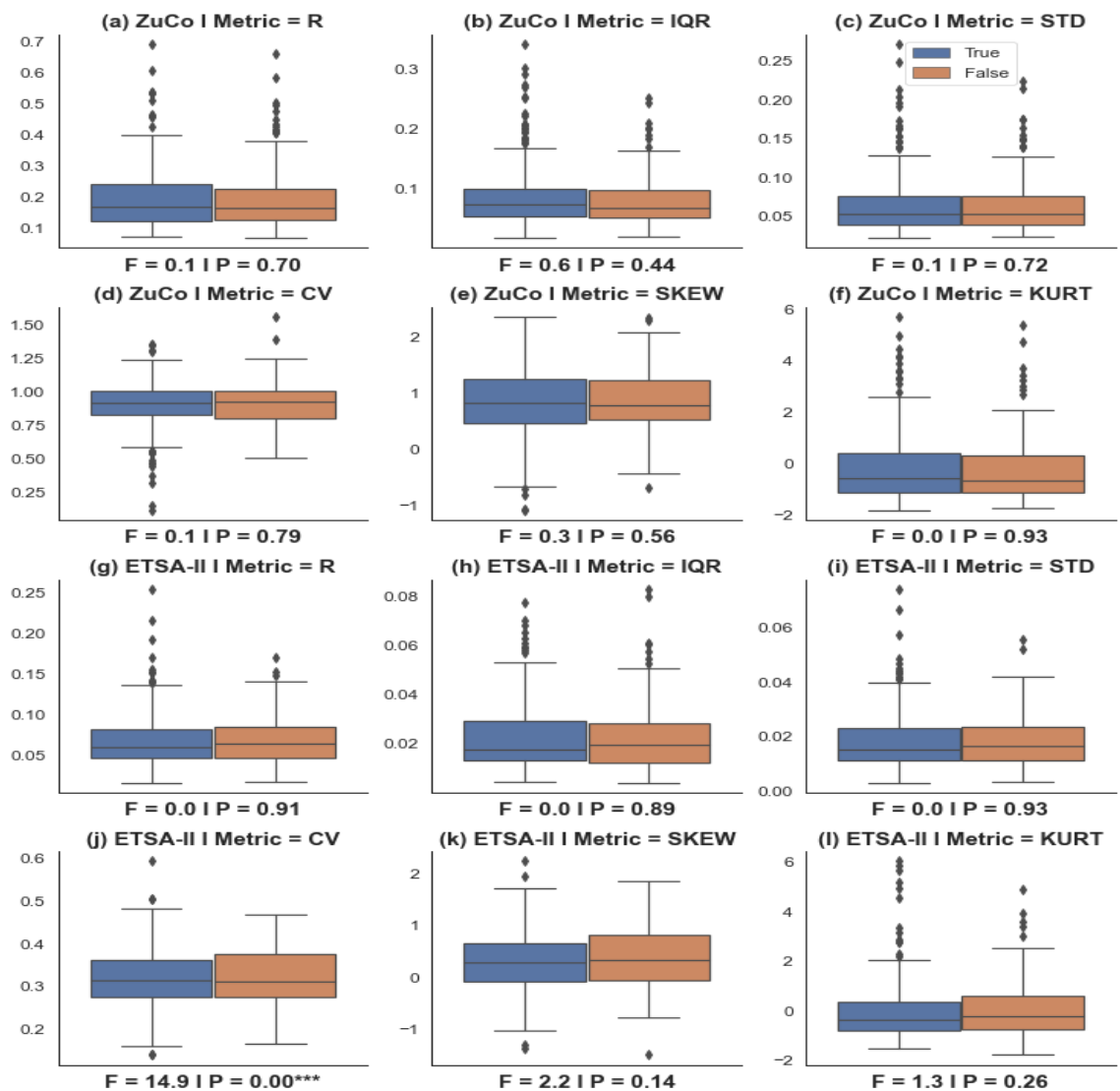

Note: ***. Significant at α=0.001.

Figure 5. The one-way ANOVA test results of machine attention values under different prediction results

***There is not a difference between the machine attention values under the correct and wrong prediction results.*** We grouped the correct and wrong prediction results, and used one-way ANOVA to investigate whether the distribution characteristics of machine attention values: range (R), interquartile distance (IQR), standard deviation (STD), coefficient of variation (CV), skew and kurt have significant differences. The results are shown in Fig. 5. The significance level α is 0.001. Except that the p-value of CV is less than 0.001 in Fig. 5(J). The p-values of other statistics are all greater than 0.001 in the other sub-figures of Fig. 5. This shows that the machine attention values does not show significant difference whether the prediction result is correct or wrong. Therefore, in the subsequent analysis, we do not distinguish whether the prediction result is correct when we analyze the machine attention values and eye-tracking values.

### 4.3 Comparison between Machine Attention values and Eye-tracking Values

The Eye-tracking values are a record of the feature of human reading, focusing on the important words in sentences when reading. So we assume that the attention mechanism can focus on important words. To verify this hypothesis, we compared the machine attention values and the eye-tracking values. Besides, to analyze whether the attention mechanism and humans really focus on the important words in sentences, we also introduced the random attention values corresponding to each sentence.

Based on the above research, we compared the machine attention values (MA), random attention values (RAN), and four eye-tracking values (ETs), i.e., nFix, FFD, TRT, and RRT. The metrics used include Spearman correlation coefficient (SCC), JS divergence (JSD), word attention overlap rate (WAOR), part-of-speech attention rate (POSAR), and sentiment word attention rate (SWAR), which performed fine-grained analysis from global to local.

### 4.3.1 Correlation Analysis Based on Spearman Correlation Coefficient and JS Divergence

Fig. 6 shows the results of Spearman correlation coefficient and JS divergence among machine attention values, eye-tracking values and random attention values. We have following findings.

***Machine attention values are not generated randomly.*** Compared with RAN, it can be seen from Fig. 6 (a) and (b) that the SCC and JSD between MA and RAN are 0.22 and 0.15 on ZuCo, respectively. From Fig. 6 (c) and (d), the SCC and JSD between MA and RAN are 0.2 and 0.07 on ESAII, respectively. It indicates that the machine attention values are different from random attention values. Previous study found that attention mechanism learns meaningful relationship between inputs and outputs (Wiegreffe and Pinter, 2020). This paper further proves that machine attention values are generated by learning rather than random. Likewise, eye-tracking values are less correlated with random attention values.

***There is a correlation between machine attention values and eye-tracking values.*** As shown from Fig. 6 (a) and (b), the machine attention values have a high correlation with the four eye-tracking values on ZuCo, especially the highest correlation with nFix and TRT, and the lowest correlation with RRT. The above findings are valid on ETSA-II, see Fig. 6 (c) and (d). It is worth noting that the SCCs of MA and ETs on ETSA-II is lower than that on ZuCo. The JSDs between MA and ETs on ETSA-II is smaller than that on ZuCo, except RRT. Our interpretation is that on ZuCo, MA and ETs can match better in word ranking but differ widely in numerical distribution, whereas the opposite is true on ETSA-II. Although there is no strong correlation between them, this cannot be ignored and deserves further analysis.

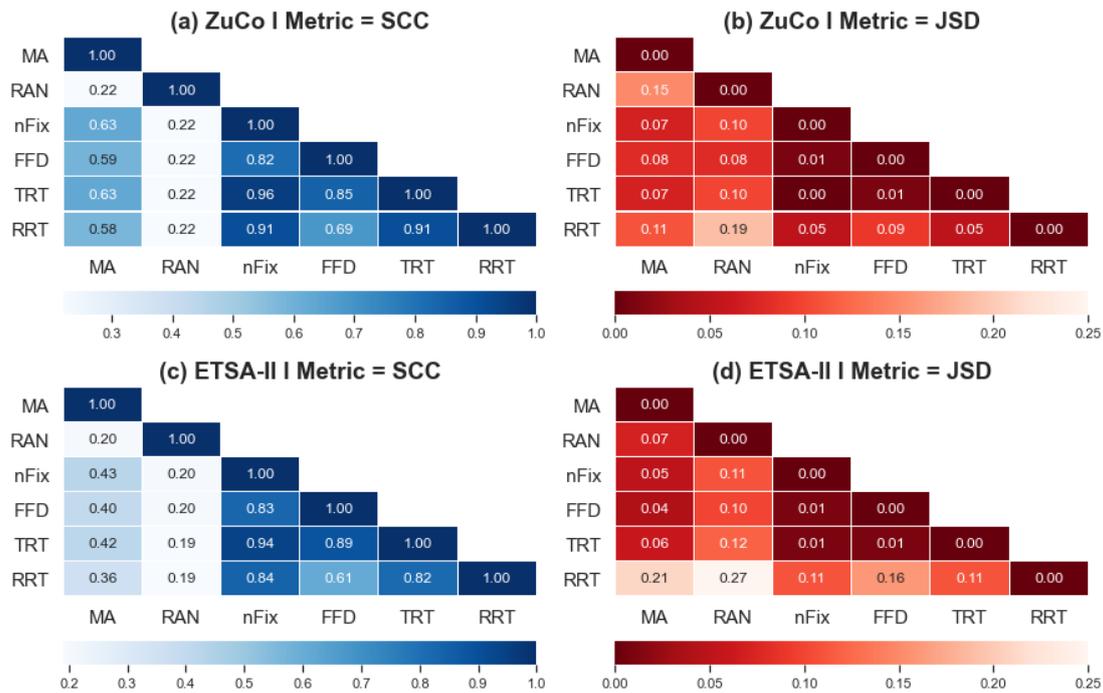

Figure 6. The heatmap of the correlation between different attention values

### 4.3.2 Whether the Attention Mechanism and Humans Focus on the Same Word

To explore whether the attention mechanism and humans focus on the same words in sentences, we used word attention overlap rate to analyze the machine attention values and the eye-tracking values. The results are shown in Fig. 7.

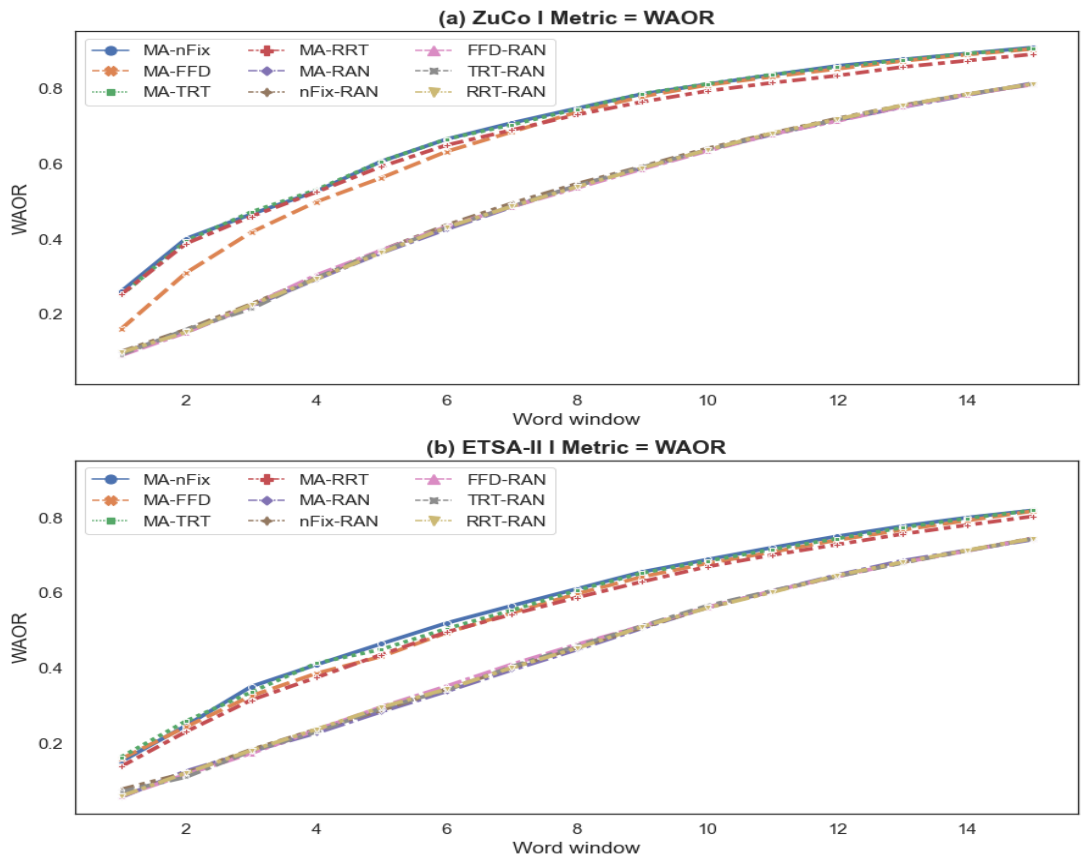

Figure 7. The trend of word attention overlap rates of different attention values

*The attention mechanism and humans can focus on the same words in sentences.* It can be seen from Fig. 7 (a) and (b) that the WAORs between MA and ETs is higher than that between MA and RAN. When the word window is 4, the attention mechanism and humans can jointly focus on about 50% of the same words on ZuCo, and about 40% of the same words on ETSA-II. It indicates that the attention mechanism and humans can focus on the same words in sentences. Specifically, in the same word window, the words considered important by the attention mechanism are within the range of words considered important by humans.

**4.3.3 Whether the Attention Mechanism and Humans Focus on Specific Words**

Based on the research in Section 4.3.2, we investigated what kind of words the attention mechanism and humans focus on. Specifically, we analyzed the attention rate of the four parts-of-speech that are valuable for expressing sentence sentiment, i.e., noun (NN), Verb (VB), adjective (JJ) and adverb (RB) (K and Devi, 2012). Besides, we used the HowNet sentiment dictionary[③] to label the sentiment words in each sentence, and calculated the attention rates of machine attention values, four eye-tracking values and random attention values on positive words (Pos) and negative words (Neg), respectively. When calculating the Part-of-speech attention rate, we set the word window to 3-5. The results are shown in Table 5 and Fig. 8.

*Adjectives are attended by the attention mechanism and humans.* From the Table 5, both the attention mechanism and humans pay more attention to adjectives in sentences on ZuCo and ETSA-II. The reason is that adjectives can usually reflect the sentiment of a sentence. On ZuCo, the ability of the attention mechanism to recognize adjectives is lower than that of humans, and the opposite is true on ETSA-II. Furthermore, on ZuCo, the attention mechanism also focuses on nouns and verbs, while on ETSA-II, the attention mechanism also focuses on adverbs. In addition to focusing on adjectives, humans also focus on nouns and adverbs on both datasets. The POSARs of RAN are obviously different from that of MA and ETs. Specifically, random attention focuses on adverbs and adjectives in sentences, but the attention to adverbs and adjectives changes dynamically under different word windows.

Table 5. The Part-of-speech attention rates of different attention values

| WIN | POS | ZuCo | | | | | | ETSA-II | | | | | |
|---|---|---|---|---|---|---|---|---|---|---|---|---|---|
| | | MA | RAN | nFix | FFD | TRT | RRT | MA | RAN | nFix | FFD | TRT | RRT |
| 3 | NN | 1.61 | 0.98 | 1.43 | 1.56 | 1.50 | 1.47 | 1.13 | 0.98 | 1.33 | 1.32 | 1.31 | 1.29 |
| | VB | 1.41 | 1.01 | 1.27 | 1.18 | 1.17 | 1.16 | 1.11 | 0.90 | 1.29 | 1.24 | 1.31 | 1.29 |
| | JJ | **1.67** | **1.13** | **2.04** | **1.83** | **2.07** | **2.07** | **2.87** | 1.10 | **1.78** | **1.79** | **1.76** | **1.66** |
| | RB | 1.11 | 1.11 | 1.53 | 1.40 | 1.49 | 1.47 | 1.97 | **1.16** | 1.32 | 1.19 | 1.35 | 1.44 |
| 4 | NN | 1.59 | 1.01 | 1.45 | 1.53 | 1.48 | 1.45 | 1.23 | 0.95 | 1.33 | 1.28 | 1.35 | 1.27 |
| | VB | 1.43 | 1.00 | 1.25 | 1.23 | 1.21 | 1.15 | 1.13 | 0.94 | 1.30 | 1.26 | 1.28 | 1.28 |
| | JJ | **1.66** | 1.03 | **1.92** | **1.73** | **1.94** | **1.93** | **2.51** | **1.10** | **1.57** | **1.68** | **1.66** | **1.50** |
| | RB | 1.05 | **1.09** | 1.47 | 1.41 | 1.47 | 1.49 | 1.90 | 1.09 | 1.30 | 1.26 | 1.27 | 1.35 |
| 5 | NN | 1.57 | 1.03 | 1.46 | 1.49 | 1.49 | 1.44 | 1.26 | 1.00 | 1.29 | 1.28 | 1.30 | 1.23 |
| | VB | 1.37 | 0.98 | 1.23 | 1.22 | 1.21 | 1.18 | 1.16 | 0.93 | 1.24 | 1.20 | 1.27 | 1.27 |
| | JJ | **1.69** | 1.03 | **1.80** | **1.68** | **1.81** | **1.80** | **2.27** | 1.02 | **1.56** | **1.63** | **1.58** | **1.48** |
| | RB | 1.14 | **1.09** | 1.46 | 1.40 | 1.39 | 1.41 | 1.85 | **1.02** | 1.32 | 1.25 | 1.31 | 1.35 |

*The attention mechanism can recognize sentiment words.* From Fig. 8, we found that under different sentiment tags, the proportion of positive words and negative words in sentences are different, as shown in the "all" item in the figure. For example, in Fig. 8 (a), when the sentiment tag is negative, the negative words in sentences account for a large proportion. In Fig.8 (c), the positive words in sentences account for a large proportion when the sentiment tag is positive. On ETSA-II, we found that when the sentiment tag is negative, the attention mechanism pays more attention to negative words than humans. When the sentiment tag is positive, the attention mechanism pays more attention to positive words than humans, as shown in Fig. 8 (d) and (e). On ZuCo, the SWARs of MA is equivalent to that of ETs. However, when the sentiment tags are neutral and positive, human attention is better than machine attention, as shown in Fig. 8 (a), (b) and (c). Compared with random attention, we found that machine attention performed better than random attention on ZuCo and ETSA-II.

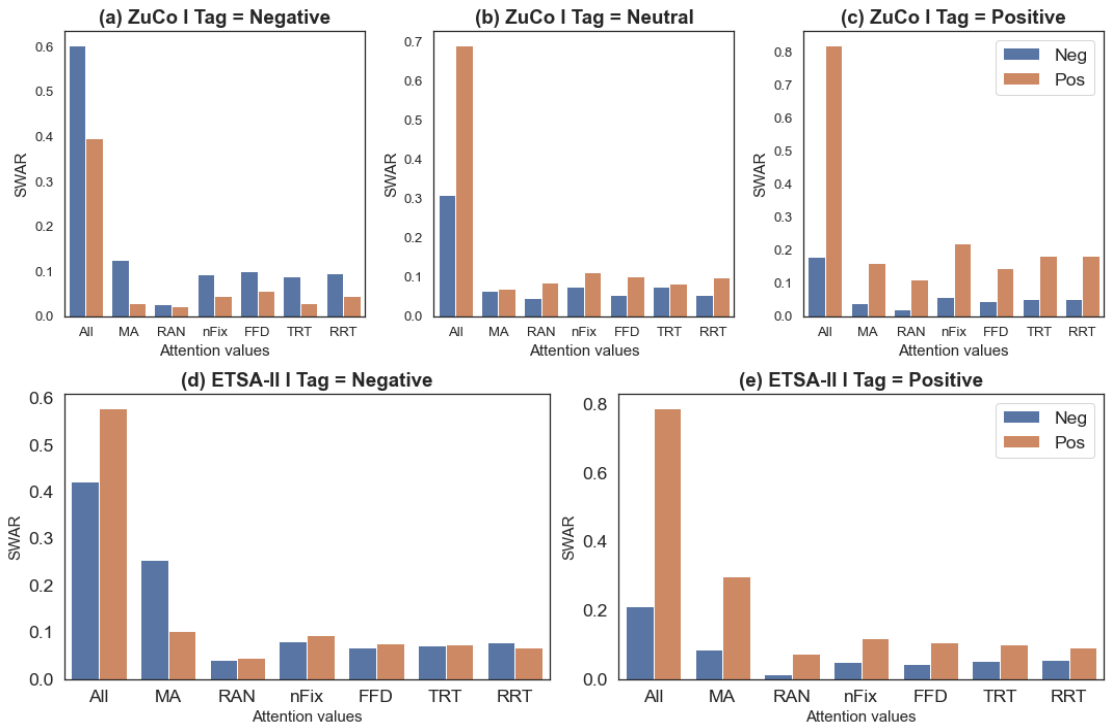

Figure 8. The sentiment word attention rates of different attention values

**4.4 The Effect of Eye-tracking Values on the Model**

By analyzing eye-tracking values and machine attention values, we found that the attention mechanism possesses the feature of human reading, focusing on important words in sentences when reading. To further reveal whether there is a correlation between machine attention and human attention, we introduced eye-tracking values into the model to enable fine-tuning of machine attention values.

*The eye-tracking values can improve model performance.* Referring to the model proposed by Mishra *et al.* (2018), we also constructed a multi-task network model, as shown in Fig. 9. The model takes the sentiment classification as the primary task and eye-tracking prediction as the auxiliary task to fine-tune the machine attention values. We conducted experiments on ZuCo and ETSA-II, and the results are shown in Table 6. We can see that the performance of the model has been improved after adding the auxiliary task. Especially, the improvement has increased significantly on ZuCo. We believed that although the attention mechanism can focus on the important words in sentences, it also has the wrong attention due to the insufficient learning. However, eye-tracking values can provide corrections for this error.

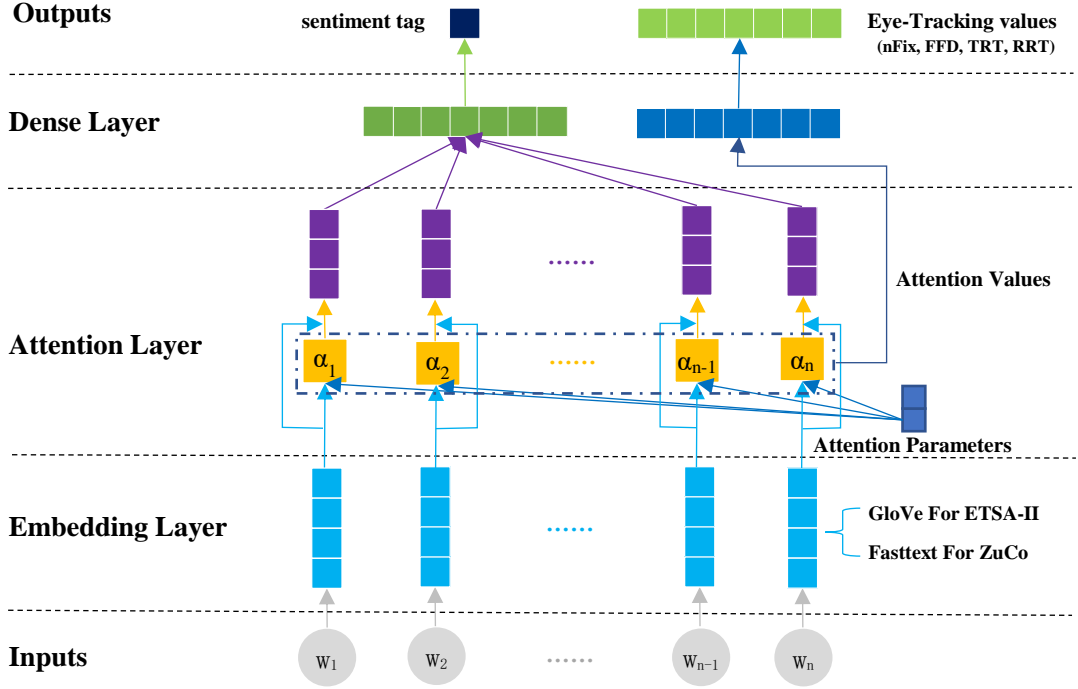

Figure 9. A multi-task model with sentiment classification as the primary task and eye-tracking prediction as the auxiliary task

Table 6. The model performance after adding eye-tracking prediction auxiliary task

|  | ZUCO | | | ETSA-II | | |
| --- | --- | --- | --- | --- | --- | --- |
|  | **P** | **R** | **F$_1$** | **P** | **R** | **F$_1$** |
| Baseline | 68.18 | 67.89 | 65.92 | 78.68 | 80.31 | 79.08 |
| +nFix | 70.06 | 70.45 | 68.71 | 79.61 | 81.12 | 80.06 |
| +FFD | 68.25 | 68.41 | 66.69 | 79.95 | 81.60 | 80.39 |
| +TRT | 69.33 | 69.75 | 68.57 | 79.56 | 81.34 | 79.94 |
| +RRT | 68.18 | 68.46 | 66.65 | 79.57 | 80.97 | 80.01 |

*The eye-tracking values can optimize machine attention.* Fig. 10 shows the correlation changes between machine attention values and eye-tracking values before and after adding eye-tracking prediction auxiliary task (MA_* represents the machine attention values obtained after adding eye-tracking prediction auxiliary task). It can be seen that the correlation between the machine attention values and the eye-tracking values is improved after adding eye-tracking prediction auxiliary task. It is worth noting that the SCCs of the fine-tuned machine attention values and eye-tracking values improved on ETSA-II, while the JSDs did not change significantly. Our interpretation is that the machine attention values before fine-tuned and ETs match better the numerical distribution, but differ greatly in word ranking, which has been explained in Section 4.3.1. However, the fine-tuned machine attention values were able to make up for this shortcoming, narrowing the difference in word ranking, and promoted the SCCs to rise, while the JSDs remains unchanged. To be more convincing, we enumerate three examples from ETSA-II, as shown in Fig. 11. We can see that the JSDs of MA and nFix, MA_nFix and nFix are equal, and the SCCs of MA_nFix and nFix is larger than that of MA and nFix. The eye-tracking values did help the attention mechanism to optimize the attention on some words weights. For example, the eye-tracking values help the attention mechanism improve the attention to the words "best" and "television" in Example 1. This improvement can also be seen in two other examples.

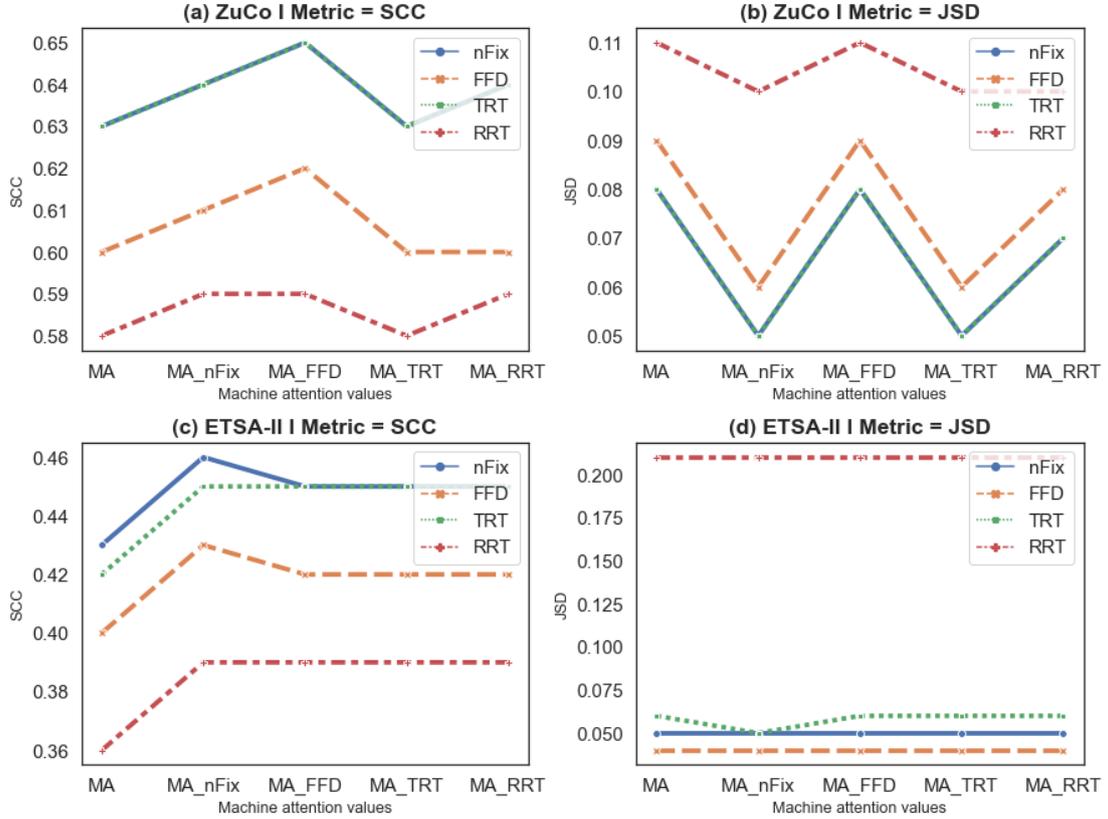

Figure 10. The correlation changes between machine attention values and eye-tracking values before and after adding the eye-tracking prediction auxiliary task

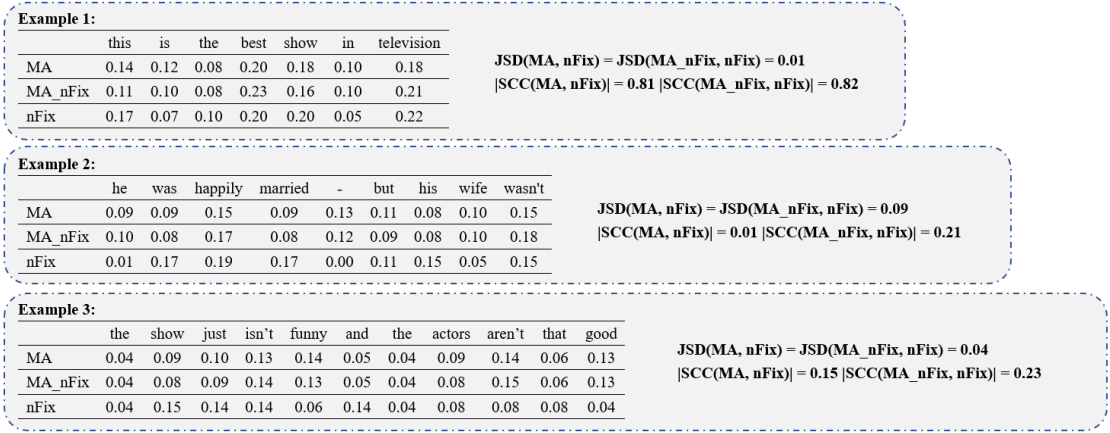

Figure 11. Three examples of eye-tracking values fine-tuning machine attention values from ETSA-II

## 5. Discussion

This study is in the line of interpretability research of the attention mechanism. We proposed to use eye-tracking values to explore the relationship between machine attention and human attention. The eye-tracking values are a record of the feature of human reading, focusing on the important words in sentences when reading. Therefore, we demonstrated that the machine attention mechanism possesses this feature of human reading. Besides, after adding the auxiliary task of eye-tracking prediction to the model, the effect of the eye-tracking values on the machine attention values is analyzed. The following will discuss whether the attention mechanism is interpretable from a human-like perspective and the value of eye-tracking values to deep learning models.

**5.1 Human-like Interpretation of Attention Mechanism**

In recent years, many studies have explored the interpretability of attention mechanism, but no unified consensus has been reached (Jain and Wallace, 2019; Lei, 2017; Mullenbach *et al.*, 2018; Sen *et al.*, 2020; Wiegreffe and Pinter, 2020). Especially, the studies of Jain and Wallace (2019) and Wiegreffe and Pinter (2020) came to opposite conclusions. By analyzing their methods, we found that by modifying the machine attention value to observe its impact on the model performance, some unforeseen errors may occur. Because it is difficult to directly observe what the attention mechanism actually learns in the model, some small changes can affect the model outputs. This has prompted some scholars to explain the role of the attention mechanism from a human-like perspective, using human attention to explain the rationality of machine attention (Sen *et al.*, 2020; Zhang, Sen, *et al.*, 2022).

Our study also explored what the attention mechanism is from a human-like perspective. We proposed to use eye-tracking values to explore the relationship between machine attention and human attention. The eye-tracking values can be regarded as a representation of human attention, as it records how long and how many times a human looks at a word in a sentence while reading. We conducted experiments on a sentiment classification task. The machine attention values were obtained form the model based on the additive attention mechanism. The reason why we use the additive attention mechanism is determined by our experimental datas. We found that the attention mechanism can focus on the important words considered by humans, and has a high degree of overlap. For a sentence, the attention mechanism can recognize adjectives and sentiment words, and even surpass human recognition ability. This indicates that the attention mechanism possesses the feature of human reading, focusing on the important words in sentences. Therefore, we believe that the attention mechanism can be explained from the human-like perspective. Although we only experiment with additive attention mechanism, we believe that more advanced attention mechanisms also possess these features. For example, Clark *et al.* (2019) claimed that the BERT's attention captures a lot of grammatical information, and can focus on the direct objects of verbs, determiners of nouns, objects of prepositions, and coreferent mentions. Htut *et al.* (2019) found that some specialist attention heads track individual dependency types in BERT and RoBERTa. Goldberg (2019) found that the attention-based BERT model can capture syntactic regularities in sentences. These studies also confirmed this conclusion.

**5.2 The Value of Eye-tracking values to Deep Learning Models**

To improve the performance of deep learning models, the study of adding eye-tracking values to the models has attracted the attention of many scholars (Agrawal and Rosa, 2020; Barrett *et al.*, 2016, 2018; Long *et al.*, 2017; Zhang and Zhang, 2021). It is generally believed that human attention can help the models correct some wrong attention, so that they can learn the features of samples better to output more accurate results (Rong *et al.*, 2021). Therefore, in our study, we also integrated the eye-tracking values into the sentiment classification model, expecting to fine-tune the machine attention values and improve the model performance.

In Section 4.4, we designed a multi-task network model, as shown in Fig. 9. The model takes the sentiment classification as the primary task and eye-tracking prediction as the auxiliary task to fine-tune the machine attention values. Through experiments, we found that the model performance was improved after adding the eye-tracking prediction task. The eye-tracking values can fine-tuning the machine attention values so that it can correctly capture important words in sentences. The correlation between the fine-tuned machine attention values and eye-tracking value is also improved. This indicates that eye-tracking values can indeed optimize machine attention. Our study provides a rationale for the studies using eye-tracking values to improve model performance. How to use the eye-tracking values in models, we have summarized in Section 2.1, and will not repeat it here.

## 6. Conclusion and Future Works

To explore the attention mechanism possesses the feature of human reading, focusing on the important words in sentences when reading, we compared the machine attention values obtained by the sentiment classification model based on additive attention mechanism and the eye-tracking values obtained by two open-source ETCs, i.e, ZuCo and ETSA-II. Through experiments, we found the attention mechanism can focus on important words, such as adjectives, adverbs, and sentiment words, which are valuable for judging the sentiment of sentences on the sentiment classification task. It indicates that the attention mechanism possesses the feature of human reading, focusing on important words in sentences when reading. Due to the insufficient learning of the attention mechanism, some words are wrongly focused. The eye-tracking values can help the attention mechanism correct this error and improve the performance of the model.

Our research yielded some interesting conclusions in simple models, and some more advanced models were dropped here. The reason is that some encoding layers affect the distribution of machine attention values, so we only consider obtaining machine attention values in a clean environment to perform the comparison with eye-tracking values. In future work, we will consider the machine attention values obtained after adding various coding layers, and analyze their comparative analysis with eye-tracking values. Besides, the current study is only designed for a sentiment classification task, and we can consider more tasks in the future to analyze the relationship between machine attention and human attention.

## Acknowledgement

This study is supported by the National Natural Science Foundation of China (Grant No. 72074113).